\documentclass[letterpaper, 10 pt, conference]{ieeeconf}  % Comment this line out
\IEEEoverridecommandlockouts                              % This command is only
\usepackage[utf8]{inputenc}
\usepackage[T1]{fontenc}
\usepackage{graphicx}
\usepackage{subcaption}
\usepackage{amsmath}
\usepackage{xcolor}
\usepackage{multirow}
\usepackage{amsfonts}
\usepackage{algorithm}% http://ctan.org/pkg/algorithms
\usepackage{algpseudocode}%http://ctan.org/pkg/algorithmicx
\usepackage{svg}
\usepackage{url}
\usepackage{paralist}
\usepackage{units}

\floatstyle{ruled}
\newfloat{algorithm}{tbp}{loa}
\providecommand{\algorithmname}{Algorithm}
\floatname{algorithm}{\protect\algorithmname}

\title{\LARGE \bf
Flow-FL: Data-Driven Federated Learning \\
for Spatio-Temporal Predictions in Multi-Robot Systems
}

\author{%
  \authorblockN{%
    Nathalie~Majcherczyk, Nishan~Srishankar
    and
    Carlo Pinciroli}
  \authorblockA{Robotics Engineering, Worcester Polytechnic Institute, MA, USA\\
  Email: \{nmajcherczyk, nsrishankar, cpinciroli\}@wpi.edu}
}

\begin{document}

\maketitle
\thispagestyle{empty}
\pagestyle{empty}

%%%%%%%%%%%%%%%%%%%%%%%%%%%%%%%%%%%%%%%%%%%%%%%%%%%%%%%%%%%%%%%%%%%%%%%%%%%%%%%%
\begin{abstract}

In this paper, we show how the Federated Learning (FL) framework enables learning collectively from distributed data in connected robot teams. This framework typically works with clients collecting data locally, updating neural network weights of their model, and sending updates to a server for aggregation into a global model. We explore the design space of FL by comparing two variants of this concept. The first variant follows the traditional FL approach in which a server aggregates the local models. In the second variant, that we call \textit{Flow-FL}, the aggregation process is serverless thanks to the use of a gossip-based shared data structure. In both variants, we use a data-driven mechanism to synchronize the learning process in which robots contribute model updates when they collect sufficient data. We validate our approach with an agent trajectory forecasting problem in a multi-agent setting.
Using a centralized implementation as a baseline, we study the effects of staggered online data collection, and variations in dataflow, number of participating robots, and time delays introduced by the decentralization of the framework in a multi-robot setting. 

%by giving periodical authorization to a robot to act as the server through a collaborative decision-making process. Results show that (we hopefully outperform the single robot training in terms of training time and generalize better to multiple environments).

% "In this paper, we propose a novel approach for
% agent motion prediction in cluttered environments. One of the
% main challenges in predicting agent motion is accounting for
% location and context-specific information". Our contribution lies in the decentralization of learning predicted trajectories from observed trajectories by a team of robots. Distributed dataset 
% towards on-line learning.

\end{abstract}

%%%%%%%%%%%%%%%%%%%%%%%%%%%%%%%%%%%%%%%%%%%%%%%%%%%%%%%%%%%%%%%%%%%%%%%%%%%%%%%%
\section{INTRODUCTION}\label{sec:introduction}

Robot swarms promise capabilities beyond the reach of single-robot solutions by distributing intelligence, sensing and actuation at a large scale~\cite{Brambilla2013}. This opportunity often comes with the challenge of dealing with large amounts of data which are physically distributed across robots. Replicating, broadcasting, and learning large-scale data has prohibitive communication and memory costs. %While work has been done in using distributed data storage where robots minimize data duplication~\cite{},

Federated Learning (FL) \cite{mcmahanarcas} is a recent approach to distributed machine learning that takes advantage of distributed data sets by partitioning learning on several machines. Data sets are partitioned either for excessive size, or because of the necessity to ensure data privacy among different data sources. In FL, individual clients collect data and calculate a local model update that is subsequently aggregated into a shared model by a dedicated server.

In a multi-robot setting, FL is a natural approach because it takes advantage of the inherently local fashion in which robots collect and process data. However, a practical approach to realizing FL in this setting is currently missing~\cite{kairouz2019advances}.

% "Swarm robotics is a branch of collective robotics that studies decentralized solutions for the problem of coordinating large teams of robots. Robot swarms are  a promising technology for large-scale scenarios, in which performing distributed tasks would entail prohibitive costs for single-robot solutions~\cite{Brambilla2013}". 
\begin{figure}[h]
    \centering
    \includegraphics[width=\linewidth]{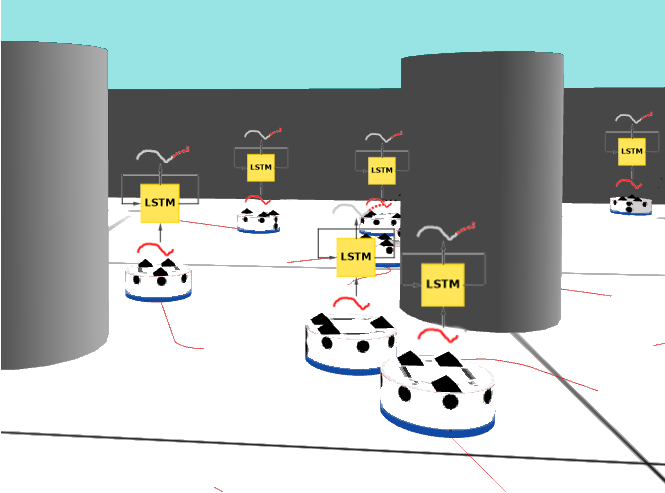}
    \caption{Federated Learning for collective trajectory forecasting in a multi-robot setting.}
    \label{fig:application}
\end{figure}

% \nm{SUPERVISED LEARNING example where it is easy to collect labelled samples.} 
In this paper, we conduct a study of the design space of FL solutions in a multi-robot setting. We compare two variants. In the first, inspired by the original FL idea, a server aggregates the model updates calculated by the robots. In the second, the server is replaced by a shared data structure.

In both cases, a central problem is how to synchronize the learning process. Data is typically collected at diverse rates across a team of robots. This affects the frequency at which robots contribute to the update of the shared model. To cope with this issue, we propose a data-driven approach in which only the robots that have collected sufficient data calculate a model update and share it with the rest of the system. Being dependent on the data flow, we named our fully distributed approach \textit{Flow-FL}.

We envision our approach to be effective in scenarios in which data varies across both time and space. To validate this insight, we consider multi-robot navigation in densely occupied environments. In these environments, the intent of other moving entities may be unknown, and the density and motion patterns might vary over time. Recent work focuses on the creation of machine learning approaches to model motion trajectories that enable predictive navigation \cite{collision_avoidance_drl}. In our approach, the robots build a shared trajectory model using locally collected data.

% We consider the case of a team of deployed robots in a environment with unknown dynamic obstacles. 

%This makes the case for adaptive navigation strategies that are able to capture changes in a timely manner and/or generalize across conditions. 

% Therefore, we propose to leverage the distributed nature of robot swarms for gathering large amounts of real-time data and performing distributed on-line learning. 

The main contributions of our paper are:
\begin{itemize}
    \item We conduct an exploration of the design space of FL in robotics settings, comparing a variant in which models are aggregated on the server with a variant in which data is aggregated in a serverless, shared data structure;
    \item We apply the proposed approaches to a trajectory prediction problem and make our open-source federated dataset available for the research community;
    \item Using a centralized implementation as a baseline, we study the effects of staggered online data collection, variations in dataflow, number of participating robots, and time delays introduced by the decentralization of the framework in a multi-robot setting. 
\end{itemize}

The paper is organized as follows. We discuss related work in Section \ref{sec:related_work} and formalize the problem in Section \ref{sec:preliminaries}. The design of our framework is presented in Section \ref{sec:methodology}.  We report the  results of  our performance evaluation in Section \ref{sec:evaluation}, and conclude the paper in Section \ref{sec:conclusion}.

% \begin{itemize}
%     \item We benchmark our Distributed Federated Learning framework against classical Machine Learning as well as Federated Learning scenarios and evaluate loss curves and trajectory prediction accuracy for multiple behaviors.
%     \item We evaluate the scalability of our system by varying the number of robots in a system, the number of data points robots need before performing a local update, and the number of robots that need to be ready before performing a global update.
%     \item We propose a novel method of decentralizing federated learning tasks by using a distributed data structure in place of a central aggregating server to store local weights or schedule updates.
% \end{itemize}

% Explore federated learning design space quorum quota parameters
% Trajectory forecasting supervised learning multi robot systems + robotics and datasets
% Decentralized 

% Generalization

\section{RELATED WORK}\label{sec:related_work}

Early attempts at distributed machine learning were based on centrally stored datasets which were processed by multiple clients~\cite{peteiro2013survey}. In these approaches, data partitioning occurs through assignment mechanisms that promote desirable features such as statistical independence and load balancing across the clients. Model updates are calculated on data that is physically copied from the server to the clients. Other early attempts applied local learning in settings in which naturally distributed data displayed analogous properties such as statistical independence and load balancing~\cite{verbraeken2020survey}.

The inception of FL was motivated by the observation that data collected in mobile devices is often not statistically independent nor necessarily balanced across clients. In addition, copying data is often undesirable due to privacy concerns.

The seminal paper on FL by McMahan \textit{et al.} \cite{mcmahanarcas} proposes \texttt{FedAvg}, which demonstrates the efficiency of averaging local model updates on a server when clients deal with non i.i.d. data and experience intermittent communication.

An important challenge in the implementation of FL is communication with the server. Particularly in distributed robotics, the presence of a server introduces the potential for a single point of failure that could endanger the success of the learning process. Therefore, decentralized approaches have surfaced which replace the server with a distributed mechanism. Notable works include using average consensus \cite{savazzi_federated_2020} and Bayesian methods \cite{lalitha2018fully, lalitha2019peer}, which trade convergence time with resilience to individual failures. The approach of George \textit{et al.} \cite{george2019distributed} offers convergence speed comparable to a server-based approach at the cost of assuming the communication topology of the clients to be fixed and predetermined.

Another important challenge is orchestrating the phases of local model update and global merging of the shared model. In traditional FL, the server takes care of this task by sending messages to the clients \cite{mcmahanarcas}. In a decentralized setting, Savazzi \textit{et al.}~\cite{savazzi_federated_2020} assume intermittent access to a centralized server and Lalitha \textit{et al.} \cite{lalitha2018fully, lalitha2019peer} allow for asynchronous communication under strong connectivity constraints.

A final key problem is selecting the clients that contribute to the model update at each learning iteration. McMahan \textit{et al.}'s research \cite{mcmahanarcas} reveals that the convergence performance of FL depends on the significance of the client updates. The significance is related to the quantity and age of the data that was used to calculate an update. Ideally, an update should be based on a large enough amount of recent data.

A recent trend in decentralizing FL is the use of conflict-free replicated data structures (CRDT)~\cite{crdt}. BAFFLE \cite{ramanan2019baffle} is an approach based on a blockchain which offers resilience to intermittent communication and the possibility to avoid global consensus at the cost of increased computational cost and local storage requirements.

Our work furthers this line of research by using a lightweight CRDT called virtual stigmergy \cite{Pinciroli2018}. In \textit{Flow-FL}, this structure is used both to schedule the learning iterations, and to store and merge the shared model.

\begin{figure*}[!t]
  \centering
   \begin{subfigure}[t]{0.3\textwidth}
        \includegraphics[width=\textwidth]{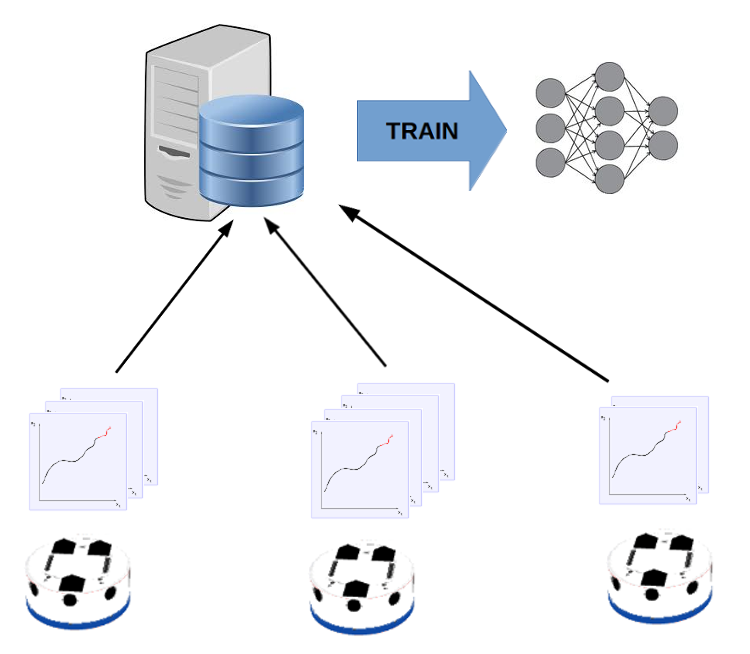}
        \caption{Centralized Machine Learning.}
        \label{sfig:centralized}
  \end{subfigure}
  \begin{subfigure}[t]{0.3\textwidth}
        \includegraphics[width=\textwidth]{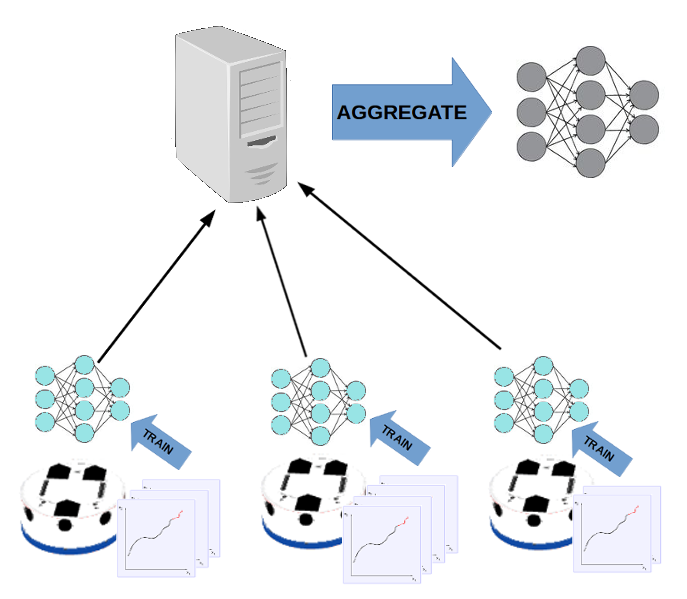}
        \caption{Federated Learning.}% DOUBLE SIDED BLACK ARROW}
        \label{sfig:FL}
  \end{subfigure}
  \begin{subfigure}[t]{0.34\textwidth}
        \includegraphics[width=\textwidth]{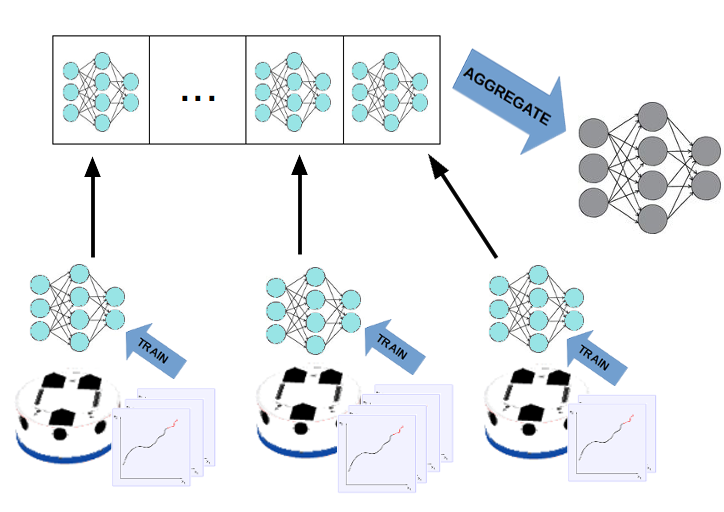}
        \caption{Distributed Federated Learning.} % DOUBLE SIDED BLACK ARROW}
        \label{sfig:DFL}
  \end{subfigure}
  \caption{Learning frameworks.}
\label{fig:partition}
\end{figure*}

\section{PRELIMINARIES}\label{sec:preliminaries}
In this section we formalize the federated learning problem and justify our application scenario.
% compare assumptions made in federated learning to our planned method.
\subsection{Federated Learning}

In machine learning, the objective function is usually of the form:
\begin{equation}
\underset{\mathbf{\Theta}}{\mathrm{min}}\thinspace L(f(x;\mathbf{\Theta}),y):= \frac{1}{n}\sum_{i=1}^{n}L(f(x^{(i)};\mathbf{\Theta}),y^{(i)})
\label{eq:ml_opt}
\end{equation}
% \iffalse
% \begin{equation}
% \underset{\mathbf{\Theta}}{\mathrm{min}}\thinspace L(f(x;\mathbf{\Theta}),y):= \frac{1}{n}\sum_{i=1}^{n}L(f(x^{(i)};\mathbf{\Theta}),y^{(i)})
% \label{eq:ml_opt}
% \end{equation}
% \fi
where the aim is to train a model $f(\cdot)$ with weights $\mathbf{\Theta}$, mapping an input $x \in \mathbb{R}^I$ to an output $y \in \mathbb{R}^O$. Using a training data set of $n$ samples $(x, y) \in \mathbb{R}^{I \times O}$, we can adjust the weights of the model to minimize a loss function $L(\cdot)$ that expresses the error between the inferred output $\hat{y}$ and the true output $y$.

\textbf{Federated Learning.}
FL~\cite{mcmahanarcas} is an optimization problem that considers a modified version of Equation~\ref{eq:ml_opt} where the training examples are now stored across $K$ clients and a global server attempts to minimize a loss function obtained from aggregated local weights (Figure~\ref{sfig:FL}). In our case, clients refer to robots that collect data. The global function that is minimized is the summation of the local losses obtained on the $k$-th robot, $L_{k}(\mathbf{\Theta})$, weighted by the number of samples observed $n_{k}$ over the total number of samples $n$ in the dataset:
\begin{equation}
\underset{\mathbf{\Theta}}{\mathrm{min}}\thinspace L(\mathbf{\Theta})=\underset{\mathbf{\Theta}}{\mathrm{min}}{\sum_{k=1}^{K}\frac{n_{k}}{n}\times L_{k}(\mathbf{\Theta})},
\label{eq:fl_opt}
\end{equation}
% \iffalse
% with $L_{k}(\mathbf{\Theta})$ being the loss obtained on the
% $k$-th robot: \fi
The local loss of the robot $L_k(\mathbf{\Theta})$ is similar to that in the traditional machine learning case, 

\begin{equation}
L_{k}(\mathbf{\Theta})=\frac{1}{n_{k}}\sum_{h=1}^{n_{k}}\ell(x_{h},y_{h};\mathbf{\Theta})\label{eq:fl_robot_loss}
\end{equation}
where $\ell(x_{h},y_{h};\mathbf{\Theta})$ is the loss of the predicted model over the $n_{k}$ samples $(x_{h},y_{h})$ observed by the robot $k$, using global model parameters $\mathbf{\Theta}$.

\textbf{Assumptions.}
FL is typically based on a number of key assumptions on the data:
\begin{inparaenum}[\it (i)]
\item It is non-i.i.d. (independent and identically distributed);
\item It is stored across several clients; and
\item It is partitioned in an imbalanced manner, resulting in clients that handle more data than others.
\end{inparaenum}
In addition, communication among clients is assumed to be intermittent. In traditional FL the presence of a server ensures synchronization, however, as discussed in Section \ref{sec:related_work}, serverless settings are also possible. In this paper, we maintain all of the above assumptions. In particular, we study the effect that the presence or absence of a server has on the performance of the learning process.

\subsection{Application: Trajectory Forecasting}

In this work we consider trajectory prediction as an application example of \textit{Flow-FL}. Trajectory forecasting is typically conducted with pedestrian data. However, existing literature on trajectory forecasting~\cite{trajnet_benchmark,RED_model,deepmap} focuses on pedestrian data collected from a single point of view which is usually an overhead camera. As such, the available datasets are not easily amenable to an FL setting. To the best of our knowledge, there are no federated datasets of trajectories collected by multiple robots. For this reason, we generated a novel federated dataset from artificial navigation data in four different multi-robot settings (see Section~\ref{ssec:datasets}). In future work, we will use real robots to collect motion data of real dynamic obstacles such as pedestrians.

Trajectory forecasting is a compelling problem for machine learning applications. This problem is about creating a model that allows robots to predict the trajectories of dynamic obstacles nearby within a short time horizon. Research has shown that navigation is significantly more efficient when using a machine learning model than with purely reactive methods, such as ORCA and RVO~\cite{rvo,orca}. While numerous neural networks have been benchmarked~\cite{trajnet_benchmark}, a simple Long Short-Term Memory (LSTM) model~\cite{lstm} has been shown to yield the lowest Average Displacement Error on the standard datasets. Therefore, we use this model architecture in our evaluation.  %Additionally, Becker \emph{et al.}~\cite{RED_model} evaluated \& justified the use of a relatively simple recurrent-encoder-dense (RED) model.

\section{METHODOLOGY}\label{sec:methodology}
% \ns{Would it make sense to have a pseud-algorithm here? ADDED AS NICE TO HAVE}
\subsection{System Design}
\label{ssec:sys_design}
\begin{figure}[t]
  \centering
  \includegraphics[width=\linewidth]{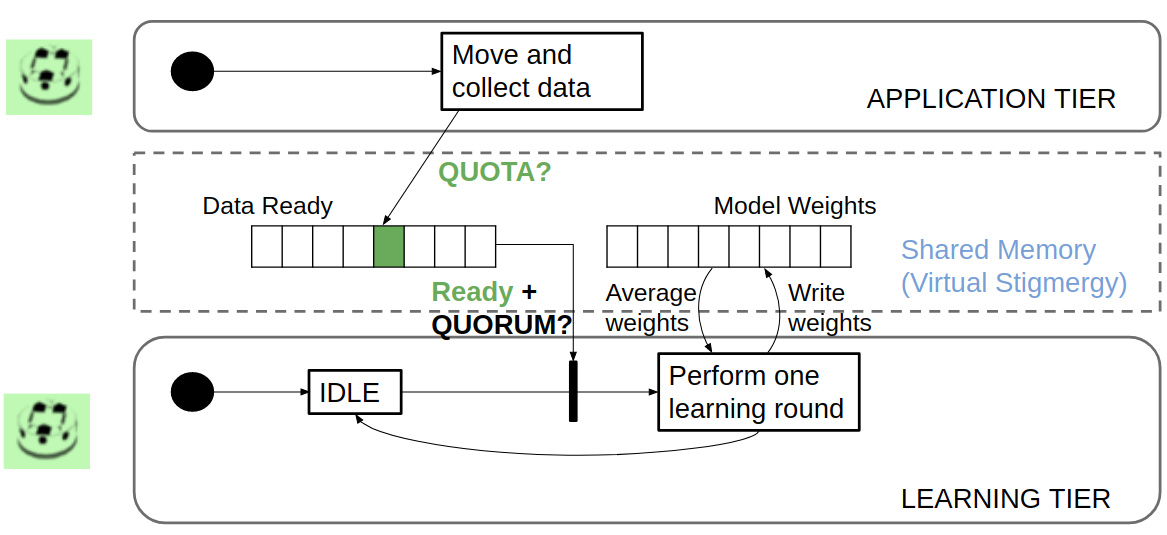}
  \caption{The state transition of the learning tier for the green robot is conditioned by the data flow of its application tier and a global state tracked in the shared memory.}
  \label{fig:state_machine}
\end{figure}

In a traditional FL setting, a server communicates with data-holding clients to enable training of a global ML model (see Figure~\ref{sfig:FL}). The server does not aggregate data, as opposed to a fully centralized approach (see Figure~\ref{sfig:centralized}). However, in traditional FL, the server has the important roles of:
\begin{inparaenum}[\it (i)]
\item orchestrating learning rounds periodically by selecting a subset of learners and sending them model parameters;
\item aggregating the results of a round of learning into a global model.
\end{inparaenum}
In our approach, depicted in Figures~\ref{sfig:DFL} and~\ref{fig:state_machine}, we replace the central server with a distributed algorithm. The scheduling of learning rounds is data-driven, and it happens when a sufficient number of robots have collected enough data. To keep the rounds sequential and distinct, a global state is maintained in a gossip-based shared memory. Model weights are written to this shared memory at the end of each round. The aggregation happens in the subsequent round, when robots with enough data pull the weights from the shared memory to instantiate their local models.

% \ns{Would it make sense to have this horizontal so it doesn' take up that much space}
% we propose a state machine gated by a distributed counter to allow periodic synchronization and we use a gossip-based shared memory to exchange data between robots.
% Each individual robot gathers data about previously seen obstacles to construct a set of sampled obstacle trajectories. The robot can then train a time series prediction model from these samples.

\textbf{Application scenario.}
We consider a scenario with $K$ moving robots that communicate and track each other's relative positions within a limited range. We tackle the problem of learning to predict the robot trajectories for a certain time horizon (see Figure~\ref{fig:application} for examples of the robot trajectories). In our setup, robots are moving, collecting data and performing learning concurrently. This is in contrast with existing frameworks for multi-robot learning, which require some sort of synchronicity in switching through these activities. To the best of our knowledge, this is an item of novelty in our framework. In our setup, the motion behavior is determined by a pre-programmed controller. We chose well-studied swarm behaviors such as flocking~\cite{flocking}, foraging~\cite{foraging}, and diffusion with obstacle avoidance~\cite{diffusion}. The data collected by the robots consists of the spatial coordinates of neighboring robots expressed in a fixed frame local to the sensing robot. A training sample is one continuous trajectory recorded for a fixed duration at regular time steps. Each robot builds its own dataset over time from its local observations.

\textbf{State machine of the learning framework (Figure~\ref{fig:state_machine}).} The operation of the robots is organized in two tiers. The learning tier runs separately from the application tier that performs the swarm behavior. %However, the learning tier depends on the data collected as input. 
The robots start in \textsc{Idle} state. Each robot has knowledge of the model architecture, but it starts with the same random weights and no data. 

The transition to the first learning round is conditioned by the flow of data. When a robot collects a certain \textit{quota} of training samples, it marks itself as ready in the shared memory. When a sufficient number of robots have marked themselves ready, we say that a \textit{quorum} has been reached and the ready robots collectively transition to the learning round. These robots, the \textit{learners}, perform local training on the data they had gathered and forget their samples. 

The next transition brings the robots back to the \textsc{Idle} state in the learning tier. The transition happens after all the learners finish sharing their new ML model weights with the rest of the swarm. Each learner shares its weights after performing one training epoch on its samples.

From now on, the transitions out of the \textsc{Idle} state proceed in a data-driven fashion as the first transition. However, the learning rounds start differently: in this case, the learners must first retrieve and aggregate the most recent weights from the shared memory.

\textbf{Shared memory.} The shared memory has two purposes: (i) It holds a global list of ready robots which is used to synchronize state transitions when the quorum of ready robots is achieved; and (ii) It stores the updated model weights in a dedicated global list.
We implemented data sharing through Virtual Stigmergy (VS)~\cite{Pinciroli2018}. VS is a lightweight, distributed tuple space designed to share a collection of \texttt{(key,value)} pairs. A local Lamport-clock-stamped copy of each tuple is stored on each robot. This copy is updated upon both \texttt{read()} and \texttt{write()} operations through network flooding. In this way, while the network topology changes, the data structure is kept up-to-date.
%The latency depends on the size of the network as it is gossip-based.

\textbf{Transition mechanism.} The state transitions are achieved through a \textit{barrier} mechanism making use of the VS. We outline the steps for this count-based consensus protocol in Algorithm~\ref{alg:barrier}.\footnote{A first implementation was proposed in \url{https://the.swarming.buzz/ICRA2017/barrier/}.}

\textbf{ML model architecture.} The ML model used for the application is a standard architecture for time series prediction. We use one Long Short-Term Memory (LSTM) layer~\cite{lstm} with a hidden dimension of 16 without returning sequences, followed by a dense layer and Dropout of \unit[0.2]{}. The network looks at a history of 32 prior $(x,y)$ observations (corresponding to \unit[3.2]{s}) and directly outputs a prediction for the next 48 time steps (\unit[4.8]{s}). We used a Mean Squared Error (MSE) loss $\text{MSE}=\frac{1}{N}\sum_{i=1}^{N}(\textbf{x}_{i}-\tilde{\textbf{x}}_{i})^{2}$ as a training metric that is then optimized using RMSProp~\cite{rmsprop_lecture}.

\textbf{ML model aggregation.} Similar to the implementation in \texttt{FedAvg}~\cite{mcmahanarcas} the \textit{learners} update the global list with their learned weights after performing a gradient descent step on local data. During the next time step, the new weights for the server are the average of all the learned weights on the list weighted by the number of samples encountered by each \textit{learner}. McMahan \emph{et al.} also found that, for large local epochs, \texttt{FedAvg} can plateau or diverge. They recommend either using fewer local epochs or decaying the number of local computations over time. We do not study this aspect, and let the \textit{learners} perform only a single local epoch per iteration and flush this data before collecting new data during the next iteration.

\begin{algorithm}[]
\caption{Barrier - robot $k$, barrier VS $\beta_k$, neighbors $\mathcal{N}_{k}$}
\label{alg:barrier} \begin{algorithmic}[1]
\Procedure{barrier\_set}{}
\State initialize barrier $\beta_k \gets list() $  for robot $k$
\State robot state $\gets$ \textbf{barrier\_wait} \normalsize
\State activate \footnotesize{\textsc{ON\_BARRIER\_UPDATE}} \normalsize
% \State$\mathbf{receive}$ 
%initialize $\mathbf{W}_{0,k}$ $\gets$ device $k$
% \For{each round $t=1,2,...$}\Comment{Main loop}
% \EndFor
\EndProcedure

\Procedure{on\_barrier\_update}{$\mathcal{N}_{k}$}
\State$\mathbf{receive}\,\,\{ i \}_{i \in \mathcal{N}_{k}}$ \Comment{RX from neighbors}
% \State $ \mathbf{merge} \, \beta_k$ and $\{\Delta \beta_i \}_{i \in \mathcal{N}_{\bar{k}}}$

\If{$i \not \in \beta_k$}%{ add $k$ to }
\State $\beta_k \gets \beta_k \cup i $
\State $\mathbf{send}\left(i \right)$ \Comment{TX to neighbors}
\EndIf
\EndProcedure

\Procedure{barrier\_ready}{$\mathcal{N}_{k}$}
\State$\mathbf{receive}\,\,\{ i \}_{i \in \mathcal{N}_{k}}$ \Comment{RX from neighbors}
\State $\beta_k \gets \beta_k \cup k $
\If {element\_count($\beta_k$) $\geq$ threshold}%
\State robot state $\gets$ \textbf{next\_state} \normalsize
\EndIf
% \For{each round $t=1,2,...$}\Comment{Main loop}
% \State$\mathbf{receive}$
% \EndFor
\State$\mathbf{send}\left(\Delta \beta_k \right)$ \Comment{TX to neighbors}
\EndProcedure

\end{algorithmic} 
\end{algorithm}

% function barrier_set() {
%   beta_k <- empty
%   robot_state <- BARRIER_WAIT
%   activate on_barrier_update
% }
% function barrier_ready() {
%   beta_k <- beta_k U k
%   if element_count(beta_k) >= threshold then
%     robot_state <- NEXT_STATE
%   endif
%   send(k)
% }
% function on_barrier_update(i) {
%   if i not in beta_k
%     beta_k <- beta_k U m_i
%     send(i)
%   endif
% }

% \clearpage

\subsection{Datasets}
\label{ssec:datasets}
To the best of our knowledge, this paper is the first to provide an open-source federated dataset of swarm motion with information on the communication graph. We generated multiple synthetic datasets of swarm motion across four distinct behaviors using ARGoS, a realistic physics-based simulator~\cite{Pinciroli:2012}.

\textbf{Data format.} Each behavior dataset %\nm{not okay with this sentence: consists of a dictionary of logged values of interest}
consists of: 

\begin{itemize}
    \item A \textit{trajectory file} that records the robot's id, neighbor id, and position across time \texttt{(robot id, neighbor id, t, x, y, z)}. Each robot records neighbor trajectories for 50,000 time steps (\unit[5,000]{s}). A trajectory sample, within a setup, is 100 time steps (\unit[10]{s}) long and is separated from other samples by an \emph{end-of-line} character. Each trajectory is expressed in the local reference frame of the robot at the start of the sample recording. This is a fixed frame of reference independent of subsequent robot motion. % This removes the effect of the motion of the robot itself.
    \item A \textit{communication graph file}, structured as \texttt{(t, robot id, neighbor id)}, that logs information about the neighboring robot IDs that are in range at every time step. This information encodes the communication graph at every time step. The communication graph file is more complete than the trajectory file because when recording the trajectory file, we drop interrupted trajectories.
    %This keeps track of the any information that is passed between robots, the density of the robot network and intermittent connectivity. \ns{Blanking here}
\end{itemize}

\textbf{Parameter setting.} The experimental settings are kept consistent across datasets as shown in Table~\ref{tab:exp_setting}. The settings include:
\begin{inparaenum}[\it (i)]
\item the trajectory sampling period;
\item additive noise for positional data on each robot neighbor sampled from a normal distribution; and
\item wheel actuation noise, also sampled from a normal distribution. The noise parameters are rounded estimates from realistic samples taken from real-world Khepera robots.
\end{inparaenum}
We executed the experiments for swarms with $K=\{15, 60\}$ robots to enable the study of the effect of different swarm sizes.

\begin{table}[!htbp]
  \centering
  \caption{Experiment settings}
  \label{tab:exp_setting}
  \begin{scriptsize}
  \begin{tabular}{l c}
  %{\linewidth}{|l @{\extracolsep{\fill}}  c|}
    \textbf{Parameter} & \textbf{Value} \\
    \hline
    %\hline
    % Number of Robots $K$ & $\{15, 60\}$ robots \\
    Trajectory duration & \unit[10]{s}\\
    Communication range & \unit[2]{m}\\
    Sensing range & \unit[2]{m} \\
    Sensing noise $n_k(t)$ & $n_k(t) \sim \unit[\mathcal{N}(0, 0.01)$]{m}\\
    Drive bias $e_k$ &  $e_k \sim \unit[\mathcal{N}(0, 0.0001)$]{m}\\
  \end{tabular}
  \end{scriptsize}
\end{table}

% rab noise 0.001
% differential steering bias_stddev=0.0001 and factor avg=1

% Fixed sample duration 

% $K = 15, 60$

% Noise on sensing: 

% Noise on differential steering:

% Sensing range:

\begin{figure}[h]
  \centering
  
  \begin{subfigure}[t]{0.2\textwidth}
        % \includesvg[width=\textwidth]{img/datasets/avoidance_15_dataset.svg}
        \includegraphics[width=\textwidth]{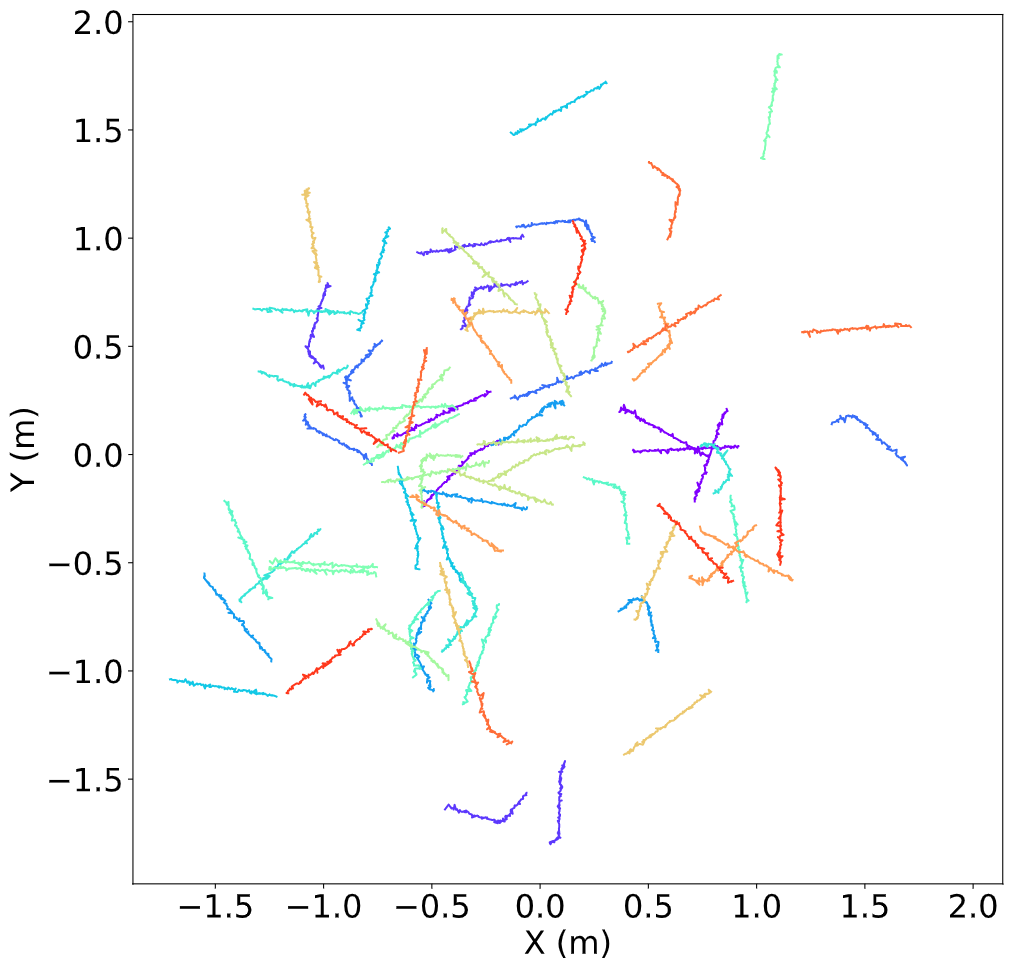}
        % \caption{Sample Obstacle avoidance trajectories with $N=15$.}
        \label{sfig:obstacleavoidance}
  \end{subfigure}
  \begin{subfigure}[t]{0.2\textwidth}
        % \includesvg[width=\textwidth,height=\textwidth]{img/datasets/avoidance_15_400traj.svg}
        \includegraphics[width=\textwidth,height=\textwidth]{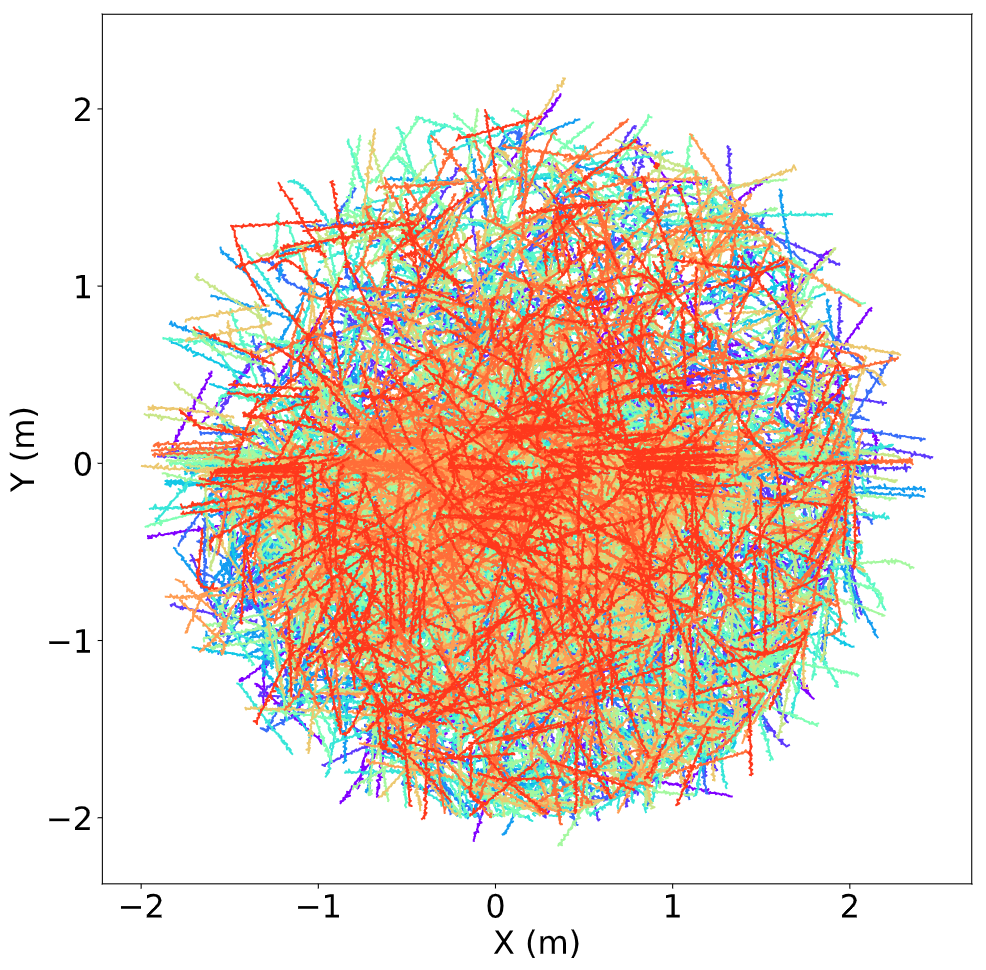}
        \label{sfig:full_avoidance}
  \end{subfigure}

   \begin{subfigure}[t]{0.2\textwidth}
        \includegraphics[width=\textwidth]{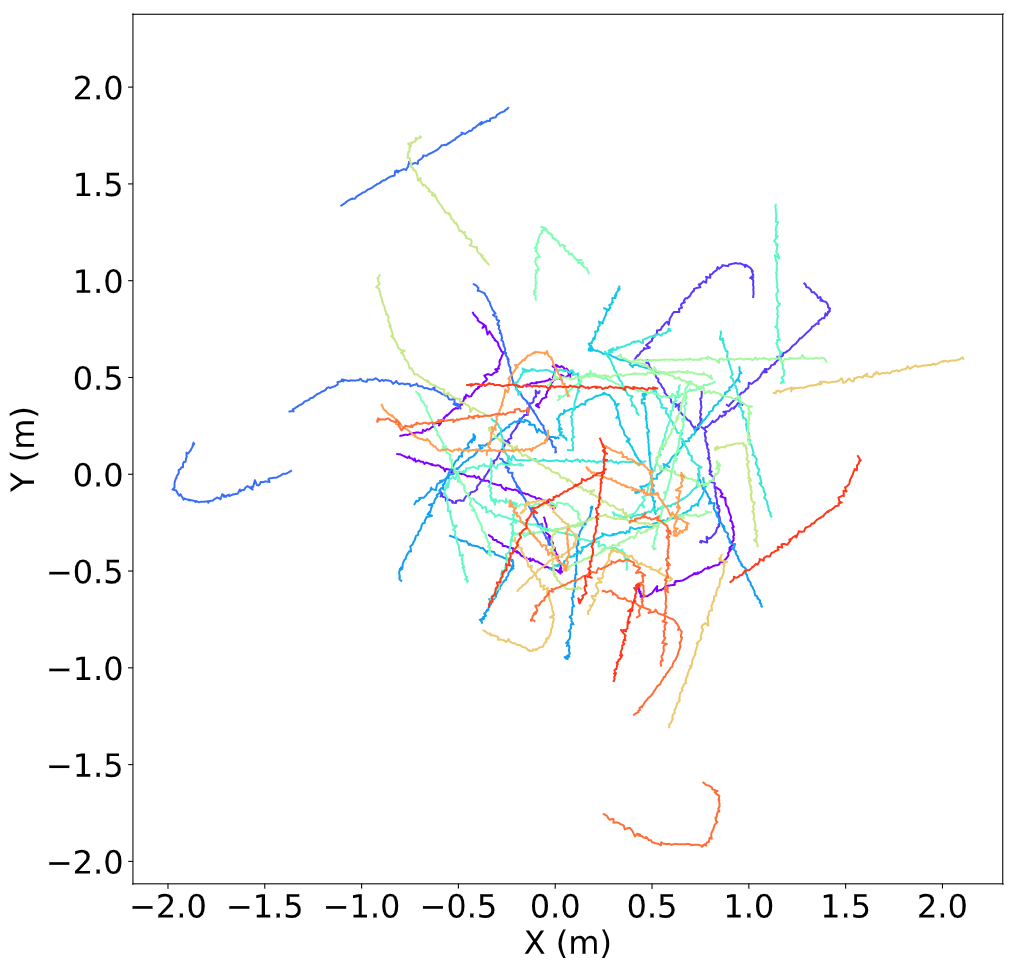}
        % \includesvg[width=\textwidth]{img/datasets/foraging_15_dataset.svg}
        % \caption{Sample foraging trajectories with $N=15$.}
        \label{sfig:foraging}
  \end{subfigure}
  \begin{subfigure}[t]{0.2\textwidth}
        \includegraphics[width=\textwidth,height=\textwidth]{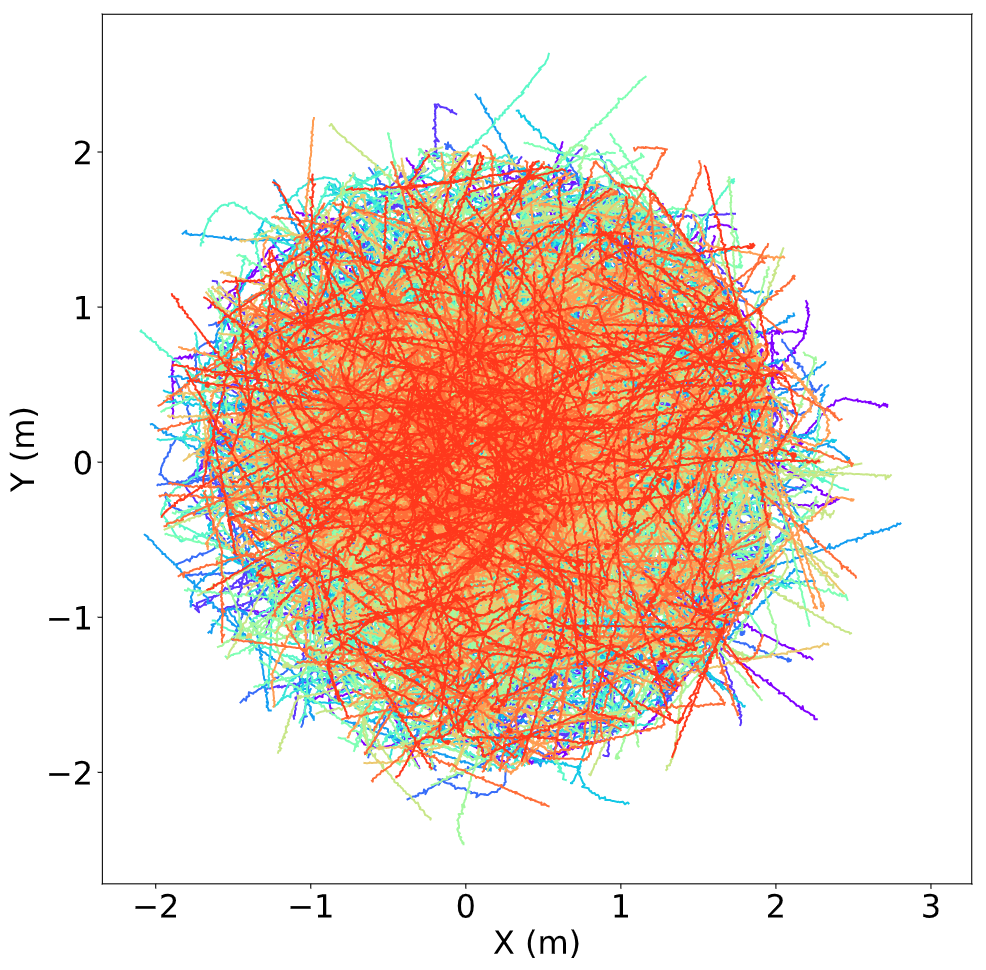}
        % \includesvg[width=\textwidth,height=\textwidth]{img/datasets/foraging_15_400traj.svg}
        \label{sfig:full_foraging}
  \end{subfigure}

  \caption{Avoidance (top) and foraging (bottom) for $K=15$. Left to right: increased number of displayed samples.}
\label{sfig:datasets}
\end{figure}

\textbf{Behavior types.} We evaluate our methodology with four different swarm behaviors~\cite{argosexamples}:
\begin{inparaenum}[\it (i)]
\item Obstacle-avoidance~\cite{diffusion} (Figure~\ref{sfig:datasets} top row) in a dense environment with uniformly distributed static obstacles;
\item Foraging~\cite{foraging} (Figure~\ref{sfig:datasets} bottom row) for resources in which robots decide whether to explore or stay in the nest according to energy considerations;
\item Phototaxis and flocking based on artificial physics~\cite{flocking}, with a light whose position is changed to prevent stagnation; and
\item A mixed behavior in which robots perform one of the previous behaviors depending on their location in the environment. 
\end{inparaenum}
The datasets and corresponding simulation videos can be found at \url{https://www.nestlab.net/doku.php/papers:mrs_fl_dataset}. Table~\ref{tab:dataset_stats} provides the total number of samples in each dataset as well as statistics about the distribution of samples between robots. The table reveals diversity across the behaviors in the number of samples collected, both total and per robot, also considering total time and 10-minute windows. In particular, the standard deviation shows how different behaviors result is different levels of imbalance in the number of samples collected by the robots.

\begin{table}[h]
\centering
\caption{Statistics for swarm motion federated datasets. }
\label{tab:dataset_stats}
\begin{tabular}{lllllll} 
\hline
Dataset & Robots & Samples & \multicolumn{2}{l}{Samples/robot} & \multicolumn{2}{l}{Samples/robot} \\
 &  &  & \multicolumn{2}{l}{} & \multicolumn{2}{c}{/10 min~} \\ 
\cline{4-7}
 &  &  & Mean & Stdev & Mean & Stdev \\ 
\cline{4-7}
\multirow{2}{*}{Avoidance} & 15 & 21,227 & 1,415 & 206 & 169 & 40 \\
 & 60 & 216,582 & 3,610 & 397 & 429 & 74 \\ 
\hline
\multirow{2}{*}{Flocking} & 15 & 49,009 & 3,267 & 427 & 390 & 78 \\
 & 60 & 333,111 & 5,552 & 436 & 659 & 112 \\ 
\hline
\multirow{2}{*}{Foraging} & 15 & 35,854 & 2,390 & 46 & 284 & 19 \\
 & 60 & 187,304 & 3,122 & 65 & 371 & 25 \\ 
\hline
\multirow{2}{*}{Mixed} & 15 & 30,627 & 2,042 & 290 & 242 & 66 \\
 & 60 & 174,863 & 2,914 & 89 & 346 & 29 \\
\hline
\end{tabular}
\end{table}

% \nm{this paragraph and comments on stats table to extended version?}
% We plotted the histogram of the x-, y- , and euclidean offsets seen in Figure~\ref{fig:datasets_hist} to explore the change in x- and y- positions from the previous time step. This shows that robots are more likely to remain stationary (e.g. when turning in place) or have incremental positional changes (within \unit[0.5]{cm}). We also note that the distributions for each of the behaviors explored are different.

% while trajectories are biased towards going straight with no $(x,y)$ offset, each of the behaviors have their own \nm{that's not exactly it, the no offset thing }

% \color{red}
% Barrier
% This behavior mimics the Quorum sensing in bacteria for collective decision making. In bacteria, it works by triggering a new behavior if chemicals reach a certain threshold. In swarm robotics, this can be done using Virtual Stigmergy to create a barrier which allows the swarm to wait for a sufficient number of robots before moving to the next bahavior. https://the.swarming.buzz/ICRA2017/barrier/
% \color{black}

\section{EVALUATION}
\label{sec:evaluation}

\subsection{Parameters of Interest}

\textbf{Quorum and quota.} The main parameters of interest in our empirical study are the \textit{quorum} of learners and the \textit{quota} of data. These parameters condition the transition to a learning state for a subset of robots at times dictated by the data collection flow (see Figure~\ref{fig:state_machine}). We set the quorum as a fraction $q_F$ of the total number of robots. We perform our study with datasets from experiments with a fixed duration of \unit[5,000]{s}. Quorum and quota determine the total number of learning rounds in each study configuration as they split the data in time. %Fixed sample length too, explain that this plays a role in time in between rounds.
% \ns{Centralized baseline epoch=round to have quorum,quota comparison} \nm{see section B}

\textbf{Number of robots.} We also study the influence of the \textit{number of robots} by using 15- and 60-robot datasets. We expect this parameter to act on different aspects of multi-robot communication and learning as it changes:
\begin{inparaenum}[\it (i)]
\item the total amount and frequency of data collected, which changes the time between learning rounds (Section~\ref{ssec:timing});
\item the robot communication network size and topology, which change the latency of the shared data structure and affect the timing of learning rounds (Section~\ref{ssec:timing});
\item the partitioning of federated data across clients, which influences the convergence rate (Section~\ref{ssec:convergence}). 
\end{inparaenum}

\textbf{Federated datasets.} We use multiple \textit{federated datasets} detailed in Section~\ref{ssec:datasets} with different behaviors (see Figure~\ref{sfig:datasets}). We provide and compare results for these 8 datasets.

\textbf{Learning hyperparameters.} We set the learning \textit{hyperparameters} for the local machine learning model as detailed in Section~\ref{ssec:sys_design}. In terms of standard FL hyperparameters, we set number of local epochs $E$ to 1. Increasing $E$ reduces communication overhead at the cost of increased individual computational load. McMahan \emph{et al.}~\cite{mcmahanarcas} provide an empirical study of the effect of this parameter. They show that high values of $E$ can lead the FL algorithm to diverge. In the same paper, McMahan \emph{et al.} also vary the quorum fraction but refer to it as client fraction. In this paper, we focus on the effect of controlling the data flow through $q_F$ and quota rather than changing the amount of local computation through $E$.

\textbf{Dataset split.} To separate our data into training/validation/test samples, we proceed in the following way: we take \unit[80]{\%} (first \unit[4,000]{s}) of the experiment data for training and validation and keep the remaining for testing; within a learning round, each learner splits the data into a set of training samples (the first \unit[80]{\%} of trajectories), and validation samples (the last \unit[20]{\%} of trajectories). We verified that behaviors do not change at the end of the experiment so as to have an appropriate testing set.

\begin{table}[!htbp]
  \centering
  \caption{Evaluation parameters}
  \label{tab:params}
  \begin{scriptsize}
  \begin{tabular}{l c}
  %{\linewidth}{|l @{\extracolsep{\fill}}  c|}
    \textbf{Parameter} & \textbf{Value} \\
    \hline
    %\hline
    Number of robots $K$ & $\{15, 60\}$ robots \\
    Quorum fraction $q_{F}$ & $\{0.2, 0.6\}$ \\
    Quorum & $q_F \cdot K$ robots\\
    Quota & $\{20, 60\}$ samples \\
    Local epochs $E$ & 1 \\
  \end{tabular}
  \end{scriptsize}
\end{table}

\subsection{Convergence Analysis}

\label{ssec:convergence}

An important aspect of our FL framework is the effect of scheduling rounds according to quorum and quota on the learning convergence. We want to study the following aspects empirically across several datasets: 
 \begin{itemize}
     \item which $(q_F,$ \textit{quota}) configuration requires the \textit{least learning rounds} to achieve convergence. Reducing the number of learning rounds reduces communication rounds thereby decreasing communication overhead; 
     \item which $(q_F,$ \textit{quota}) configuration requires data spread across the \textit{least time steps} to achieve convergence. Reducing the number of steps gives us a final model earlier on in the experiment;
     \item which $(q_F,$ \textit{quota}) configuration gives us the \textit{best trade-off} between the two above situations.
 \end{itemize}
 
Varying quorum and quota is effectively re-partitioning the data in time within the same dataset. With respect to learning, this affects the rate of model updates as well as the number of participating clients $\bar{K}$ and the number of local examples $n_{k}$ used for the update at each round. 

We study the effect of $(q_F$, \textit{quota}) on the federated validation loss $L(\mathbf{\Theta} )$ from Equation~\ref{eq:fl_opt}. To make the comparison fair with \textit{Flow-FL}, we implemented a data-driven FL approach with a server scheduling rounds according to $q_F$ and \textit{quota}, i.e., starting a round as soon as $q_F \cdot K$ robots have a \textit{quota} of samples. In the distributed version Flow-FL, the learning rounds are scheduled the same way, but they occur with a delay after the quorum/quota condition is met. This delay is due to the latency imposed by the update of the shared data structure. Thus, a different number of learners may qualify at the same time. To compare convergence in a consistent way, we define the stopping round as the round where the windowed average of loss changes less than a threshold of $0.0001$. The averaging window was set to 5 rounds.% (one order of magnitude lower than loss value).

\begin{figure*}[h!]
  \centering
  \begin{subfigure}[t]{0.49\textwidth}
      \includegraphics[width=\textwidth]{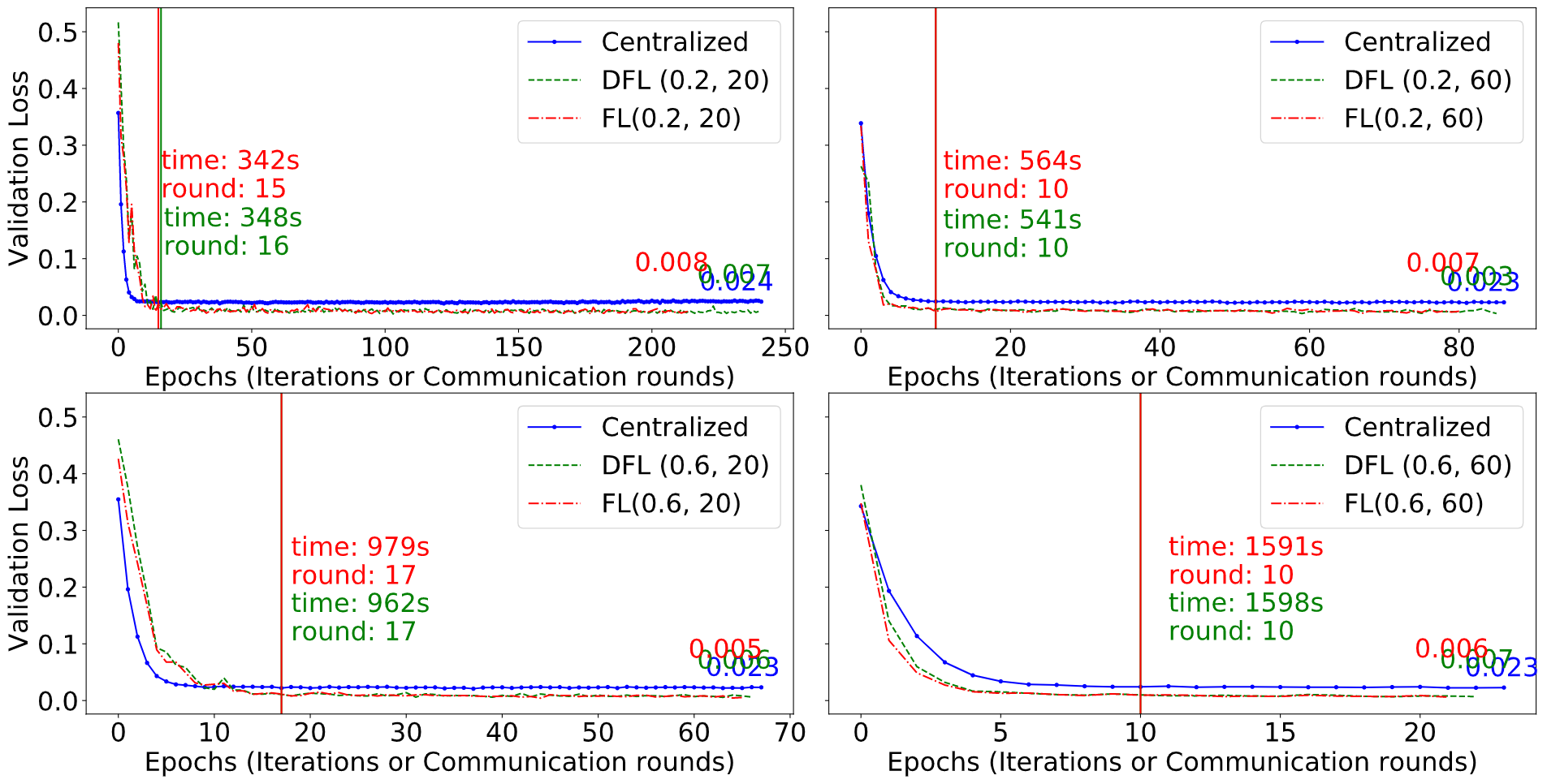}
    %   \label{fig:validation15}
  \end{subfigure}
\begin{subfigure}[t]{0.49\textwidth}
    \includegraphics[width=\textwidth]{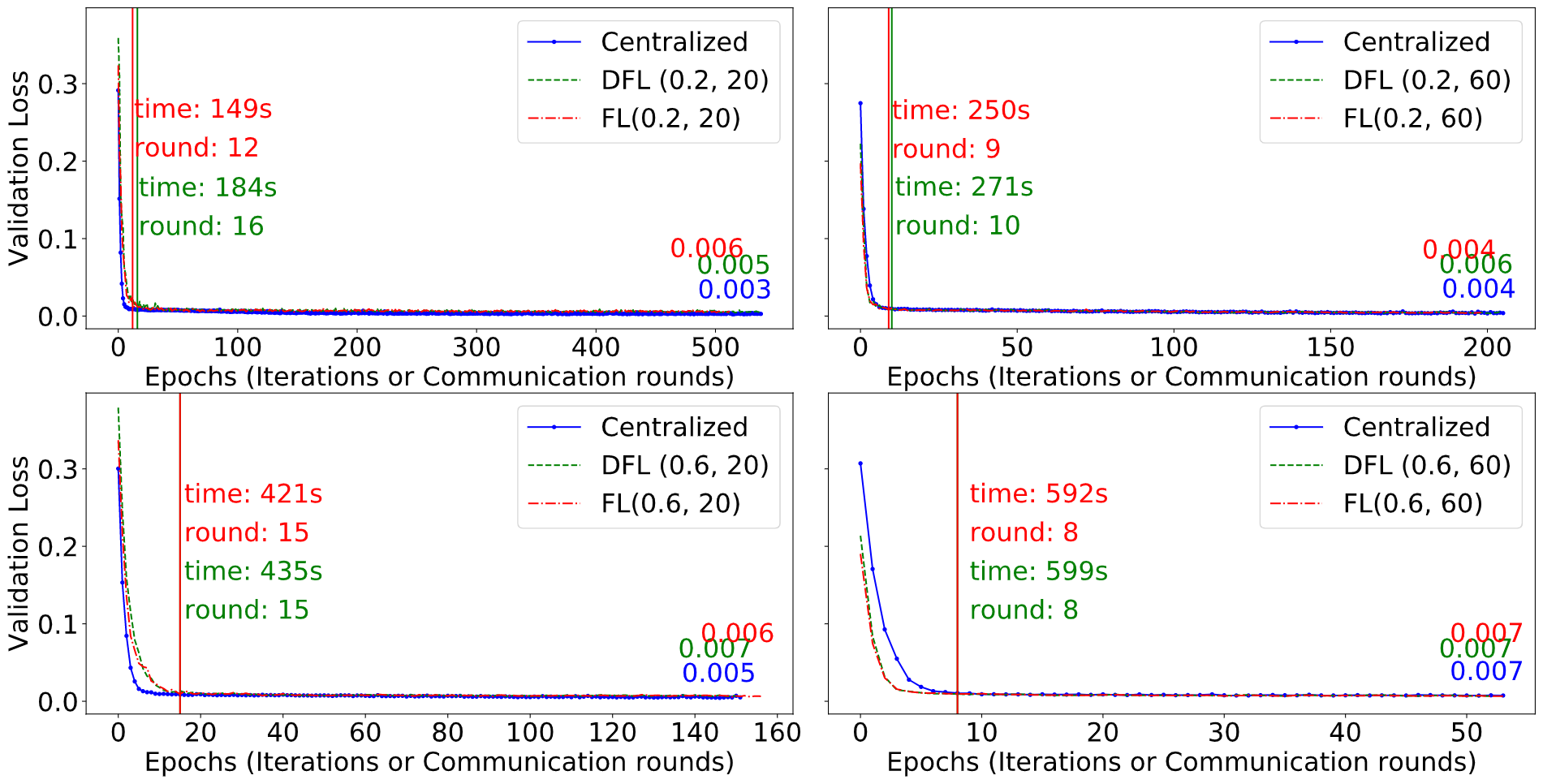}
    
  \end{subfigure}
\caption{Validation loss $L(\mathbf{\Theta})$ for the avoidance behavior with 15 (left), 60 (right) robots (DFL=\textit{Flow-FL}).}
\label{fig:validation}
\end{figure*}

\textbf{Discussion.}
Figure~\ref{fig:validation} shows the federated validation loss curves. We also show the centralized validation loss as a baseline that uses all the data collected by the end of the experiment at once. To compare the loss over iterations, we show epochs for the centralized loss and learning rounds for the federated loss. We include as many epochs for the baseline as the number of learning rounds in \textit{Flow-FL}. Table~\ref{tab:loss_behaviors60} reports the final loss value and stopping points for all the behaviors. The final loss is similar across configurations, but it tends to increase as the total number of iterations decreases. With higher $(q_F$, \textit{quota}), we have fewer learning rounds because it takes more data to move to the next training round. We also note fewer oscillations of the learning curve with higher $(q_F$, \textit{quota}).

\textbf{Stopping rounds and times.} Table~\ref{tab:loss_behaviors60} shows the lowest stopping rounds and times across configurations in bold:
\begin{itemize}
    \item  We see that, while changing across behaviors, the lowest number of learning rounds occurs with higher thresholds of ($q_F$, \textit{quota}) than the minimum $(0.2, 20)$;
    \item  We get lower numbers of time steps at stopping with lower thresholds for ($q_F$, \textit{quota}). However, the stopping criterion sometimes selected an early stopping round with some oscillations occurring later in the curve. Those instances are denoted by an asterisk. 
    \item The best trade-off between number of rounds and time steps is at $(q_F$, \textit{quota}) $= (0.2, 60)$.
\end{itemize}

% Note: Loss provided for centralized at end of simulation is different between runs with quorum/quota because different number of epochs. Made to match the Flow-FL training rounds.
%Different loss in FL versus DFL because 

\begin{table}
\centering
\caption{Convergence data across quorums and quotas for different behaviors ($K=60$).}
\label{tab:loss_behaviors60}
\begin{tabular}{llllll}
 \textbf{($q_F$, quota)}  &  & \textbf{(0.2 20)}  & \textbf{(0.2, 60)}  & \textbf{(0.6, 20)}  & \textbf{(0.6, 60)}  \\ 
\hline\hline
Flocking &  &  &  &  &  \\ 
\cline{1-1}
Validation loss & C & 0.002 & 0.002 & 0.002 & 0.002 \\
 & FL & 0.002 & 0.001 & 0.001 & 0.001 \\
 & Flow-FL & 0.002 & 0.002 & 0.002 & 0.002 \\ 
\hline
Stopping round & FL & 17 & 8 & 16 & \textbf{6} \\
 & Flow-FL & 17 & \textbf{8} & 13 & \textbf{8} \\ 
\hline
Stopping time(s) & FL & \textbf{134*} & 157 & 297 & 292 \\
 & Flow-FL & \textbf{136*} & 157 & 247 & 402 \\ 
\hline\hline
Foraging &  &  &  &  &  \\ 
\cline{1-1}
Validation loss & C & 0.011 & 0.011 & 0.012 & 0.016 \\
 & FL & 0.014 & 0.013 & 0.017 & 0.014 \\
 & Flow-FL & 0.012 & 0.013 & 0.016 & 0.015 \\ 
\hline
Stopping round & FL & 21 & \textbf{7} & 13 & 8 \\
 & Flow-FL & 15 & 10 & 16 & \textbf{7} \\ 
\hline
Stopping time(s) & FL & \textbf{257} & \textbf{257} & 448 & 731 \\
 & Flow-FL & \textbf{192} & 326 & 562 & 636 \\ 
\hline\hline
Avoidance &  &  &  &  &  \\ 
\cline{1-1}
Validation loss & C & 0.003 & 0.004 & 0.005 & 0.007 \\
 & FL & 0.006 & 0.004 & 0.006 & 0.007 \\
 & Flow-FL & 0.005 & 0.006 & 0.007 & 0.007 \\ 
\hline
Stopping round & FL & 12 & 9 & 15 & \textbf{8} \\
 & Flow-FL & 16 & 10 & 15 & \textbf{8} \\ 
\hline
Stopping time(s) & FL & \textbf{149*} & 250 & 421 & 592 \\
 & Flow-FL & \textbf{184*} & 271 & 435 & 599 \\ 
\hline\hline
Mixed &  &  &  &  &  \\ 
\cline{1-1}
Validation loss & C & 0.006 & 0.007 & 0.009 & 0.012 \\
 & FL & 0.008 & 0.008 & 0.010 & 0.010 \\
 & Flow-FL & 0.010 & 0.009 & 0.012 & 0.011 \\ 
\hline
Stopping round & FL & 20 & \textbf{8} & 19 & \textbf{8} \\
 & Flow-FL & 16 & \textbf{8} & 15 & \textbf{8} \\ 
\hline
Stopping time(s) & FL & \textbf{232} & 242 & 611 & 705 \\
 & Flow-FL & \textbf{192} & 241 & 503 & 711
\end{tabular}
\end{table}

\subsection{Learning Round Timing}
\label{ssec:timing}

\begin{figure}[h!]
    \centering
    \includegraphics[width=\linewidth]{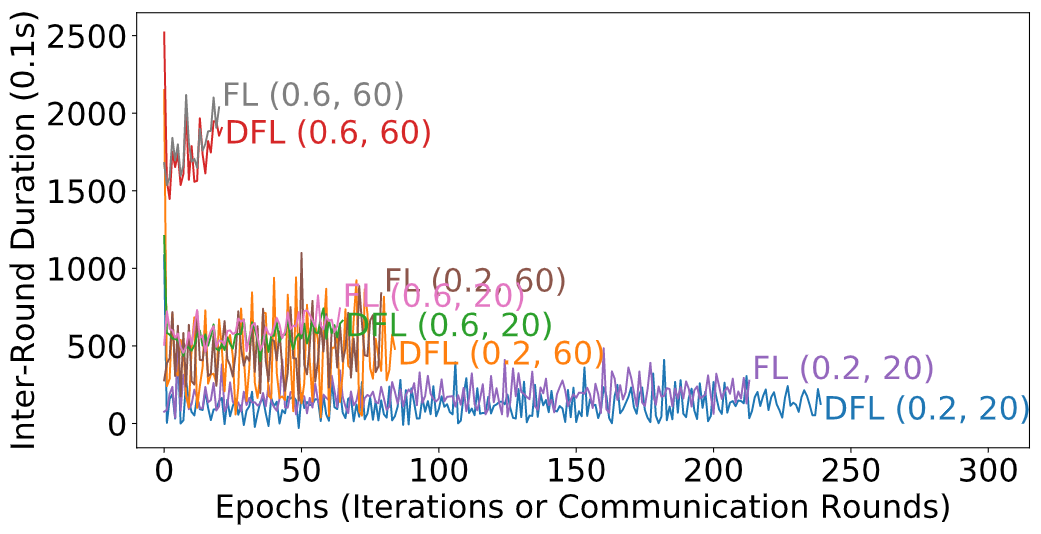}
    \caption{Timing for avoidance with 15 robots (DFL=\textit{Flow-FL}).}
    \label{fig:round_timing}
\end{figure}

The scheduling of learning rounds is data-driven both in our version of FL and in \textit{Flow-FL}. The time between learning rounds increases when we increase either the learner quorum or the sample quota. Figure~\ref{fig:round_timing} shows the time between rounds over the duration of the experiment. This time depends on the flow of data in the experiment and the length of the time series we consider as samples. The graph shows that, for a given (quorum,quota) setting, the inter-round duration is similar between FL and \textit{Flow-FL}. This is due to the short time spent by the robots in the barrier with \textit{Flow-FL}. With $K=15$, the average time spent in the \textit{barrier} is 15 time steps (\unit[1.5]{s}) and, with $K=60$, it is 51 time steps (\unit[5.1]{s}). A wider empirical study of the latency of virtual stigmergy is reported in~\cite{Pinciroli2018}.

\subsection{Prediction Quality}
We quantify the prediction quality of trajectory forecasting based on two metrics: 
\begin{inparaenum}[\it (i)]
\item The Average Displacement Error (ADE) in Equation~\ref{eq:ade}, that is a summation of the L2-norm between the ground truth $\mathbf{y_{i,t}}$ and the predicted trajectory $\mathbf{\tilde{y}_{i,t}}$ over the predicted horizon $T$ for all trajectory samples $N_\text{traj}$;
\item The Final Displacement Error (FDE) in Equation~\ref{eq:fde} which is the summation of the L2-norm of the final positions between the predicted $\mathbf{\tilde{y}_{i,T}}$ and the ground truth $\mathbf{y_{i,T}}$ over all trajectory samples.
\end{inparaenum}

\begin{equation}
\text{ADE}=\frac{1}{N_\text{traj}\cdot~T}\sum_{i=1}^{N_\text{traj}}\sum_{t=1}^{T}\Vert(\mathbf{y}_{i,t}-\mathbf{\tilde{y}}_{i,t})\Vert
\label{eq:ade}
\end{equation}

\begin{equation}
\text{FDE}=\frac{1}{N_\text{traj}}\sum_{i=1}^{N_\text{traj}}\Vert(\mathbf{y}_{i,t=T}-\mathbf{\tilde{y}}_{i,t=T})\Vert
\label{eq:fde}
\end{equation}

\begin{table}[!htbp]
  \centering
  \caption{Trajectory reconstruction across quorums and quotas for different behaviors. ($K=60$)}
  \label{tab:evaltraj60}
%   \begin{scriptsize}
  \scalebox{0.8}{
  \begin{tabular}{llllll}
%   \textbf{(Quorum, quota)} & & \textbf{(0.2 20)} & \textbf{(0.2, 60)} & \textbf{(0.6, 20)} & \textbf{(0.6, 60)} \\
    $\mathbf{(q_F,}$ & & \textbf{(0.2,} & \textbf{(0.2,} & \textbf{(0.6,} & \textbf{(0.6,} \\
    \textbf{quota)} & & \textbf{20)} & \textbf{60)} & \textbf{20)} & \textbf{60)} \\
    \hline
    \hline
    \multirow{1}{*}{\shortstack[r]{Flocking}}
    \\
    \cline{1-1}
    \unit[FDE]{(m)}  
    & Centralized & 0.08$\pm$0.04 & 0.08$\pm$0.04 & 0.08$\pm$0.04 & 0.09$\pm$0.04\\
    & FL & 0.08$\pm$0.04 & 0.08$\pm$0.04 & 0.09$\pm$0.05 & 0.08$\pm$0.04\\
    & Flow-FL & 0.09$\pm$0.05 & 0.08$\pm$0.04 & 0.09$\pm$0.05 & 0.09$\pm$0.04\\
    \hline
    \unit[ADE]{(m)} 
    & Centralized & 0.05$\pm$0.03 & 0.05$\pm$0.03 & 0.05$\pm$0.03 & 0.05$\pm$0.03\\
    & FL & 0.05$\pm$0.03 & 0.05$\pm$0.03 & 0.05$\pm$0.03 & 0.05$\pm$0.03\\
    & Flow-FL & 0.06$\pm$0.03 & 0.05$\pm$0.03 & 0.06$\pm$0.03 & 0.05$\pm$0.03\\
    \hline
    \hline
    \multirow{1}{*}{\shortstack[r]{Foraging}}
    \\
    \cline{1-1}
    \unit[FDE]{(m)}  
    & Centralized & 0.22$\pm$0.11 & 0.23$\pm$0.11 & 0.23$\pm$0.11 & 0.27$\pm$0.13\\
    & FL & 0.25$\pm$0.12 & 0.23$\pm$0.12 & 0.27$\pm$0.13 & 0.25$\pm$0.12\\
    & Flow-FL & 0.24$\pm$0.12 & 0.23$\pm$0.12 & 0.27$\pm$0.13 & 0.25$\pm$0.12\\
    \hline
    \unit[ADE]{(m)} 
    & Centralized & 0.11$\pm$0.05 & 0.12$\pm$0.05 & 0.12$\pm$0.06 & 0.14$\pm$0.06\\
    & FL & 0.13$\pm$0.06 & 0.12$\pm$0.06 & 0.14$\pm$0.07 & 0.13$\pm$0.06\\
    & Flow-FL & 0.13$\pm$0.06 & 0.12$\pm$0.06 & 0.14$\pm$0.07 & 0.13$\pm$0.06\\
    \hline 
    \hline
    \multirow{1}{*}{\shortstack[r]{Avoidance}}
    \\
    \cline{1-1}
    \unit[FDE]{(m)}  
    & Centralized & 0.11$\pm$0.05 & 0.13$\pm$0.05 & 0.15$\pm$0.06 & 0.19$\pm$0.06\\
    & FL & 0.17$\pm$0.06 & 0.14$\pm$0.05 & 0.19$\pm$0.06 & 0.18$\pm$0.06\\
    & Flow-FL & 0.16$\pm$0.06 & 0.14$\pm$0.05 & 0.19$\pm$0.06 & 0.18$\pm$0.07\\
    \hline
    \unit[ADE]{(m)} 
    & Centralized & 0.06$\pm$0.02 & 0.07$\pm$0.03 & 0.08$\pm$0.03 & 0.10$\pm$0.04\\
    & FL & 0.09$\pm$0.04 & 0.07$\pm$0.03 & 0.10$\pm$0.04 & 0.10$\pm$0.04\\
    & Flow-FL & 0.08$\pm$0.03 & 0.07$\pm$0.03 & 0.10$\pm$0.04 & 0.10$\pm$0.04\\
    \hline
    \hline
    \multirow{1}{*}{\shortstack[r]{Mixed}}
    \\
    \cline{1-1}
    \unit[FDE]{(m)}   
    & Centralized & 0.15$\pm$0.10 & 0.17$\pm$0.11 & 0.18$\pm$0.12 & 0.21$\pm$0.15\\
    & FL & 0.21$\pm$0.14 & 0.18$\pm$0.13 & 0.24$\pm$0.17 & 0.21$\pm$0.15\\
    & Flow-FL & 0.20$\pm$0.14 & 0.18$\pm$0.12 & 0.24$\pm$0.17 & 0.21$\pm$0.14\\
    \hline
    \unit[ADE]{(m)} 
    & Centralized & 0.08$\pm$0.05 & 0.09$\pm$0.05 & 0.10$\pm$0.06 & 0.11$\pm$0.07\\
    & FL & 0.11$\pm$0.07 & 0.09$\pm$0.06 & 0.13$\pm$0.09 & 0.11$\pm$0.07\\
    & Flow-FL & 0.11$\pm$0.07 & 0.09$\pm$0.06 & 0.13$\pm$0.09 & 0.11$\pm$0.07\\
  \end{tabular}}
% \end{scriptsize}
\end{table}

\begin{figure}[h!]

  \centering
  
%   \begin{subfigure}[t]{0.5\textwidth}
        % \includegraphics[width=\textwidth,height=\textwidth]{img/trajplots/allrobots_50trajectories/avoidance_K15/DFL_02_20.eps}
    \includegraphics[width=0.8\linewidth]{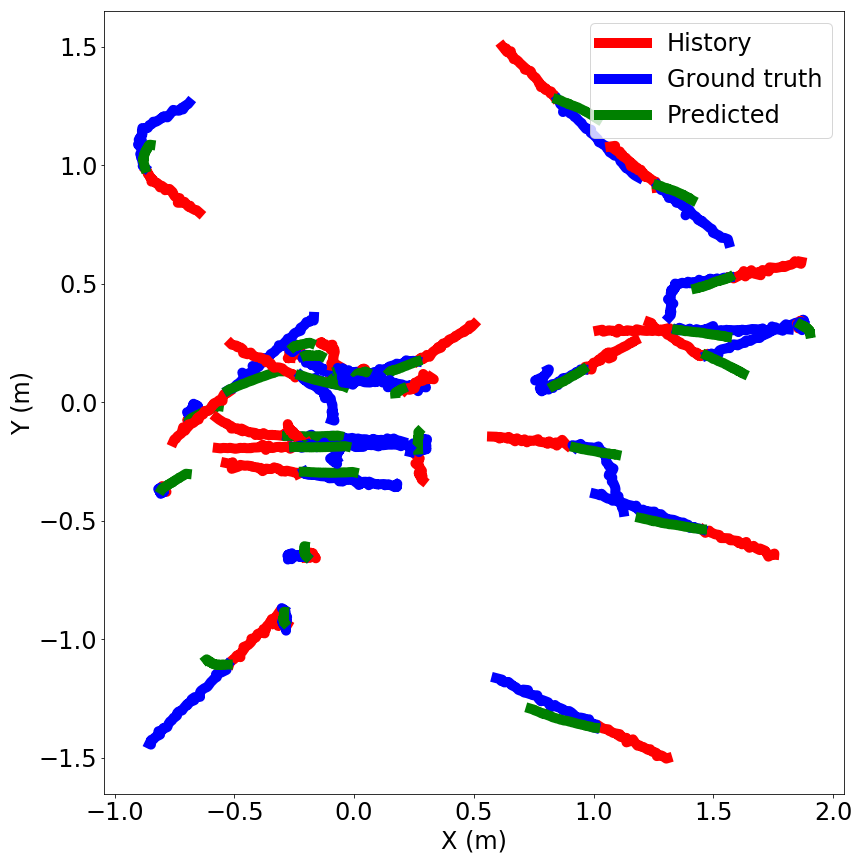}
        % \caption{Generated Obstacle avoidance trajectories with $N=15$.}
    \caption{Fifteen trajectories predicted using \textit{Flow-FL} for one robot with mixed behaviors in a swarm for $K=60$.}
     \label{sfig:gen_trajs}
        % \label{sfig:gen_mixed}
\end{figure}

We report the ADE, FDE metric results in Table~\ref{tab:evaltraj60} to 2 decimal places since this corresponds to centimeter-level granularity. We maintain consistency across the evaluations in the case of centralized training by setting the number of epochs trained to the number of rounds performed by \textit{Flow-FL} for the (quorum, quota) pair.
We note that trajectories are forecast similarly by the three methods across all four behaviors with no notable differences. We identify that, in general, reconstructing flocking trajectories which are goal-oriented to a light source is much easier than reconstructing foraging or avoidance trajectories which need additional context information from neighbors (such as distance between neighbors in the case of obstacle avoidance, or location of resources/the nest for foraging). The generated output trajectories are shown in Figure~\ref{sfig:gen_trajs}. %We notice that while the model can infer on trajectories, it does not factor contextual information. 
Data such as a robot's motion model, higher-derivative information (such as velocity), or a particular agent's goals would help robustly predicting turning or in some cases matching the distance traveled by a robot (even though the orientations are predicted accurately). These improvements are out of the scope of our work.

% \ns{Comparison to K60 in process}

% \nm{Accuracy and a few qualitative curves representing predictions}

\section{CONCLUSIONS}\label{sec:conclusion}

In this work, we explore the design space of Federated Learning in a robotics setting. Our study includes two versions of Federated Learning (one server-dependent and one serverless), and one centralized approach as our baseline. We provide a practical realization of fully distributed Federated Learning, \textit{Flow-FL}, in a multi-robot setting. We propose a way to schedule model updates based on data flow by considering two parameters: the \textit{quota}, i.e., the minimum number of data samples for a robot to qualify for a model update, and the \textit{quorum}, i.e., the minimum number of robots to start a learning round. %In robotics, this is particularly relevant because data has temporal patterns and we need to capture enough diverse samples to train machine learning models. Our scheduling also reduces imbalance in number of samples used in learning rounds by different clients.   %In order to capture enough diverse samples
%, pertinent to schedule updates with a large enough frequency.

We studied the role of several parameters of practical relevance, such as staggered online data collection, number of participating robots, and time delays introduced by decentralization. As we envision our approach to be useful in learning dynamic spatio-temporal datasets, we considered a well-known case study with compelling dynamics in both space and time: trajectory forecasting. Due to the lack of usable datasets in the literature, we created the first federated dataset from artificial data collected from a representative set of multi-robot behaviors.

In future work, we will apply \textit{Flow-FL} to real-world pedestrian data collected with a team of robots. In addition, we will study the role of the communication topology in keeping communication delays limited when the number of robots increases. Finally, we will study the role of aggressive communication loss on the convergence of \textit{Flow-FL}.

%we will characterize and improve communication efficiency, for instance  and propose data compression methods for model updates.  

%The setting of decentralized SGD also naturally lends itself to asynchronous algorithms in which each
%client becomes active independently at random times, removing the need for global synchronization and
%potentially improving scalability

% An orthogonal question for fully decentralized learning is how it can be practically realized.

% \nm{Not the same for us but I like this conclusion. Us -> empirical study and distributed communication framework.}
% \color{red}
% In this work, we have proposed FedProx, an optimization framework that tackles the systems and statistical heterogeneity inherent in federated networks. FedProx allows for variable amounts of work to be performed locally across devices, and relies on a proximal term to help stabilize the method. We provide the first convergence results of FedProx in realistic federated settings under a device dissimilarity assumption, while also accounting for practical issues such as stragglers. Our empirical evaluation across a suite of federated datasets has validated our theoretical analysis and demonstrated that the FedProx framework can significantly improve the convergence behavior of federated learning in realistic heterogeneous networks
% \color{black}

\addtolength{\textheight}{-12cm}   % This command serves to balance the column lengths

%%%%%%%%%%%%%%%%%%%%%%%%%%%%%%%%%%%%%%%%%%%%%%%%%%%%%%%%%%%%%%%%%%%%%%%%%%%%%%%%

%%%%%%%%%%%%%%%%%%%%%%%%%%%%%%%%%%%%%%%%%%%%%%%%%%%%%%%%%%%%%%%%%%%%%%%%%%%%%%%%

%%%%%%%%%%%%%%%%%%%%%%%%%%%%%%%%%%%%%%%%%%%%%%%%%%%%%%%%%%%%%%%%%%%%%%%%%%%%%%%%

\section*{ACKNOWLEDGMENT}

This work was funded by a grant from Amazon Robotics. This research was performed using computational resources supported by the Academic $\&$ Research Computing group at Worcester Polytechnic Institute.
% mRobot Technology Co, Shanghai, China.

%%%%%%%%%%%%%%%%%%%%%%%%%%%%%%%%%%%%%%%%%%%%%%%%%%%%%%%%%%%%%%%%%%%%%%%%%%%%%%%%

% \begin{thebibliography}{99}
\bibliographystyle{IEEEtran}
\bibliography{biblio}

\end{document}